\documentclass{article}
\usepackage{spconf,amsmath,graphicx,hyperref}
\usepackage{amssymb}
\usepackage{booktabs}
\usepackage{multirow}
\usepackage[dvipsnames,table]{xcolor}
\usepackage{colortbl}

\definecolor{LightCyan}{RGB}{220,230,241}

\title{Meow: End-to-End Outline Writing for Automatic Academic Survey}
\name{%
    Zhaoyu Ma$^{1}$$^{2*}$%
    \qquad Yuan Shan$^{1}$$^{3*}$\thanks{* Equally contribution, work done during internship at Wenge}%
    \qquad Jiahao Zhao$^{1}$$^{4 \dagger}$\thanks{$\dagger$ Corresponding author}
    \qquad Nan Xu$^{1}$
    \qquad Lei Wang$^{1}$
}
\address{    $^1$Beijing Wenge Technology Co.,Ltd\hspace{1cm}\\
    $^2$Peking University\hspace{1cm}$^3$Duke University\hspace{1cm}\\
    $^4$Institute of Automation, Chinese Academy of Sciences}
\begin{document}
%\ninept
%
\maketitle
\begin{abstract}
As academic paper publication numbers grow exponentially, conducting in-depth surveys with LLMs automatically has become an inevitable trend. Outline writing, which aims to systematically organize related works, is critical for automated survey generation. Yet existing automatic survey methods treat outline writing as mere workflow steps in the overall pipeline. Such template-based workflows produce outlines that lack in-depth understanding of the survey topic and fine-grained styles.
To address these limitations, we propose \textbf{Meow}, the first \textbf{me}tadata-driven \textbf{o}utline \textbf{w}riting framework that produces organized and faithful outlines efficiently. Specifically, we first formulate outline writing as an end-to-end task that generates hierarchical structured outlines from paper metadata. We then curate a high-quality dataset of surveys from arXiv, bioRxiv, and medRxiv, and establish systematic evaluation metrics for outline quality assessment. Finally, we employ a two-stage training approach combining supervised fine-tuning and reinforcement learning. Our 8B reasoning model demonstrates strong performance with high structural fidelity and stylistic coherence.
\end{abstract}
\section{Introduction}
\label{sec:intro}
The exponential growth in academic publications has made conducting comprehensive literature reviews increasingly challenging and time-consuming~\cite{doi:10.1021/prechem.5c00051}. Simultaneously, significant advances in Large Language Models (LLMs) have opened new possibilities for automated survey generation~\cite{jin2025lorvp,liang2025surveyxacademicsurveyautomation}. Within this automation pipeline, outline generation serves as the critical foundation that determines the structural and logical framework of the entire survey, requiring deep domain understanding and systematic knowledge organization.

Automatic survey generation methods can be grouped into retrieval-driven and interactive paradigms~\cite{jin2025lorvp,liang2025surveyxacademicsurveyautomation,qiu2025completingsystematicreviewhours,wang2024autosurveylargelanguagemodels,wen2025interactivesurveyllmbasedpersonalizedinteractive}. Retrieval-driven systems (e.g., AutoSurvey~\cite{wang2024autosurveylargelanguagemodels}, SurveyX~\cite{liang2025surveyxacademicsurveyautomation}) emphasize paper collection and retrieval-based writing, and interactive frameworks (e.g., InteractiveSurvey~\cite{wen2025interactivesurveyllmbasedpersonalizedinteractive}, InsightAgent~\cite{qiu2025completingsystematicreviewhours}) leverage human–AI collaboration. Yet, these agent-based approaches are inefficient due to multi-step reasoning and generate rigid outlines with shallow coherence. Related to taxonomy generation, early pattern-based and distributional methods had poor coverage, while neural approaches introduced contrastive or reinforcement learning~\cite{MENG2024112405,10.1016/j.engappai.2021.104501,meng-etal-2025-tef,sarhan-etal-2024-taxocritic}. 
% More recently, LLMs enabled end-to-end ontology learning~\cite{lippolis2025ontologygenerationusinglarge,Doumanas2025FineTuningLL} and domain workflows~\cite{fathallah2024llms4lifelargelanguagemodels}, but challenges remain for fine-grained, research-oriented taxonomies. 
Overall, outline generation still faces several technical challenges: (1) high task complexity requiring fine-grained understanding, organization, and integration of domain concepts; (2) improving factuality and ensuring deeper, more systematic structure and organization; (3) lack of high-quality datasets and systematic evaluation frameworks.

\begin{figure}[t]
\centering
\includegraphics[width=0.8\linewidth]{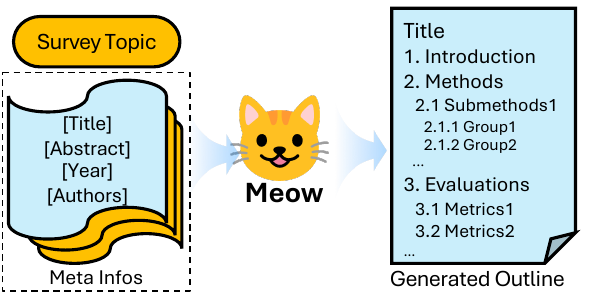}
\caption{Illustration of metadata to survey outline.}
\label{fig:task}
\vspace{-0.3cm}
\end{figure}

To address these challenges, we propose \textbf{Meow}, the first \textbf{me}tadata-driven \textbf{o}utline \textbf{w}riting framework. As shown in Figure \ref{fig:task}, we formulate survey outline generation as an end-to-end process that takes candidate paper metadata as input and produces complete hierarchical outlines in a single step, effectively leveraging LLMs' long-context comprehension capabilities. To support this approach, we construct a high-quality dataset from 2.82 million arXiv papers, supplemented with biomedical surveys from bioRxiv and medRxiv, processed into standardized format containing metadata, outline structures, and comprehensive bibliographic metadata. We also develop a systematic evaluation framework that assesses structural, pragmatic, and content-related aspects under the LLM-as-a-Judge paradigm.

Methodologically, we employ a two-stage training strategy combining Chain of Thought (CoT) cold-start with reinforcement learning. The CoT stage guides structured reasoning through theme categorization and outline development, using DeepSeek-R1~\cite{deepseekai2025deepseekr1incentivizingreasoningcapability} to generate high-quality training data. The reinforcement learning stage incorporates a structural distance reward, and format compliance to enhance both authenticity and structural rationality.

Experimental results show that our approach outperforms agent-based methods in outline generation, delivering superior structural fidelity and stylistic coherence. This validates the effectiveness of end-to-end long-context modeling. Unlike approaches focused solely on generation quality, our method incorporates structured reasoning and human-aligned reward modeling, producing outlines that better conform to academic writing conventions and standards.

We summarize our contributions as follows: (1) We formulate long-context survey outline generation in an end-to-end manner, and construct a high-quality dataset alongside a systematic evaluation framework.
(2) We design a two-stage training approach for outline generation, incorporating CoT cold-start and reinforcement learning.
(3) We experimentally demonstrate the effectiveness of the proposed method, offering a new technical pathway for automatic survey generation.

\section{Related Work}\label{sec:related}
Automatic survey generation has emerged as a critical research area with approaches categorized into retrieval-based methods like AutoSurvey~\cite{wang2024autosurveylargelanguagemodels}, SurveyX~\cite{liang2025surveyxacademicsurveyautomation}, and SurveyGo~\cite{jin2025lorvp}, and interactive frameworks like InteractiveSurvey~\cite{wen2025interactivesurveyllmbasedpersonalizedinteractive} and InsightAgent~\cite{qiu2025completingsystematicreviewhours}. Recent specialized work includes LitLLMs~\cite{agarwal2025litllmsllmsliteraturereview} for related work generation and CogWriter~\cite{wan2025cognitivewritingperspectiveconstrained} applying cognitive writing theory. In taxonomy generation, early pattern-based methods~\cite{wang-etal-2017-short} have evolved into neural approaches using contrastive learning~\cite{MENG2024112405}, reinforcement learning~\cite{10.1016/j.engappai.2021.104501,sarhan-etal-2024-taxocritic,meng-etal-2025-tef}, and taxonomy completion techniques~\cite{10.1145/3589334.3645584,TaxoRankConstruct,Peng_2024}. Large language models have enabled end-to-end ontology learning~\cite{lippolis2025ontologygenerationusinglarge,Doumanas2025FineTuningLL} with domain-specific applications~\cite{fathallah2024llms4lifelargelanguagemodels}. However, existing systems lack sophisticated mechanisms for creating semantically coherent taxonomies that accurately reflect domain-specific knowledge organization for research-oriented surveys.

\section{Proposed Method}

We propose an end-to-end framework for generating hierarchical survey outlines directly from paper metadata. Our complete framework consists of two main stages: (1) data curation from multiple academic sources with CoT annotation, and (2) two-stage training with CoT cold-start followed by reinforcement learning optimization.

\subsection{Task Definition}

We formulate hierarchical outline generation as a single-step inference process that maps paper metadata to structured academic outlines. Given a target topic $\mathcal{T}$ and a collection of $N$ papers $\mathcal{P} = \{p_1, p_2, \ldots, p_N\}$, where each paper $p_i$ contains metadata $(t_i, a_i)$ with $t_i$ as paper title and $a_i$ as paper abstract, our goal is to generate a hierarchical outline $\mathcal{O} = \{s_1, s_2, \ldots, s_M\}$ where each section $s_j$ comprises section heading $h_j$. The generated outline should maintain structural coherence while ensuring citation accuracy with respect to the input paper collection.

\subsection{Survey Data Curation} 

To create a high-quality training dataset, we curate a comprehensive collection of survey papers along with their metadata and extract human-written hierarchical outlines. Our data processing pipeline is as follows:

\textbf{Data collection}
We first collect 2.82 million metadata of papers from arXiv\footnote{https://www.kaggle.com/datasets/Cornell-University/arxiv}, along with 252k from bioRxiv and 72k from medRxiv, respectively. Details are shown in Table \ref{tab:data_sources}.

\textbf{Data filtering}
We employed a multi-stage filtering process to ensure data quality. Initially, we identified survey articles by filtering for keywords such as "survey", "review", "overview", and "meta-analysis". We then applied structural criteria to select well-organized papers with appropriate heading hierarchies and section distributions. Finally, we enforced reference integrity by requiring that all cited references in the outline appear in the bibliography, ensuring complete citation information. Non-essential sections such as acknowledgments and appendices were removed to prevent potential bias. This systematic filtering process yielded a high-quality dataset suitable for our research objectives.

\textbf{Data completion}
To enhance the model's comprehension of the references, we enriched the reference entries by supplementing them with abstracts. Specifically, we vectorized the title of each reference using the \texttt{all-MiniLM-L6-v2} model. Based on vector similarity, we then retrieved the corresponding abstracts from the metadata of our collection of 2.82 million papers and attached them to the reference entries. If no match was found, the abstract field was left empty.

\begin{table}[t]
  \centering
  \footnotesize
  \setlength{\tabcolsep}{4pt}
  \renewcommand{\arraystretch}{1.2}
  \begin{tabular}{lrr} % 第一列左对齐，数值列右对齐
    \toprule
    \textbf{Source} & \textbf{\# Total metadata} & \textbf{\# Survey} \\
    \midrule
    arXiv & 2,828,998 & 16,201 \\
    bioRxiv & 252,357 & 1,945 \\
    medRxiv & 72,059 & 5,683 \\
    \bottomrule
  \end{tabular}
  \caption{Summary of data sources and survey article counts.}
  \label{tab:data_sources}
\end{table}

\textbf{CoT distillation}
Chain of Thought(CoT) breaks down complex reasoning into systematic, step-by-step processes. In our distillation process, \texttt{DeepSeek-R1} was prompted to derive a taxonomy from the references through clustering, thereby constructing the logical deduction chain that bridges the input and final answer. The generated CoT allows the model to engage in explicit reasoning before producing output. 

\subsection{CoT Cold-Start with SFT}

We initialize the model through supervised fine-tuning on a curated CoT dataset. Given input $(\mathcal{T}, \mathcal{P})$ and target outline $\mathcal{O}^*$, the SFT objective minimizes the negative log-likelihood:

\begin{equation}
\mathcal{L}_{\text{SFT}}(\boldsymbol{\theta}) = -\mathbb{E}_{(\mathcal{T}, \mathcal{P}, \mathcal{O}^*) \sim \mathcal{D}} \left[ \sum_{j=1}^{M} \log \pi_{\boldsymbol{\theta}}(s_j | \mathcal{T}, \mathcal{P}, \mathcal{O}_{<j}) \right]
\end{equation}
where $\mathcal{D}$ is training distribution, $\boldsymbol{\theta}$ denotes model parameters, and $\mathcal{O}_{<j}$ represents previously generated sections.

\subsection{Reinforcement Learning Phase}

Following cold-start initialization, we employ Group Relative Policy Optimization (GRPO) to optimize outline quality through reward-based learning.

For each input query $q = (\mathcal{T}, \mathcal{P})$, we sample $G$ candidate outlines $\{\mathcal{O}_i\}_{i=1}^G$ from the current policy. The GRPO objective combines clipped policy updates with KL regularization:

\begin{equation}
\begin{split}
\mathcal{L}_{GRPO}(\theta) &= \mathbb{E}_{\tau \sim \mathcal{D}} \bigg[ \frac{1}{G} \sum_{i=1}^{G} \Big( \min(r_i(\theta) \hat{A}_i, \\
&\quad \text{clip}(r_i(\theta), 1-\epsilon, 1+\epsilon) \hat{A}_i) \Big) - \beta D_{KL}(\pi_\theta \| \pi_{\text{ref}}) \bigg]
\end{split}
\end{equation}
where $r_{i}(\boldsymbol{\theta}) = \frac{\pi_{\boldsymbol{\theta}}(\mathcal{O}_i | q)}{\pi_{\boldsymbol{\theta}_{\text{old}}}(\mathcal{O}_i | q)}$ is the importance ratio, $\hat{A}_i = \frac{R_i - \text{mean}(\{R_j\}_{j=1}^G)}{\text{std}(\{R_j\}_{j=1}^G)}$ is the normalized advantage estimate, $R_i$ represents the total reward for outline $\mathcal{O}_i$, $\epsilon$ controls clipping range and $\beta$ balances KL penalty strength.

\subsection{Reward Function Design}
\label{sec:reward}

We design two reward functions to ensure structural quality and format compliance.

\textbf{Structural Similarity Reward} We extract outlines as tree structures and compute similarity using Tree Edit Distance (TED). For generated outline $\mathcal{O}_{\text{gen}}$ and reference outline $\mathcal{O}_{\text{ref}}$:

\begin{equation}
R_{\text{struct}}(\mathcal{O}_{\text{gen}}, \mathcal{O}_{\text{ref}}) = 1 - \frac{\text{TED}(\mathcal{O}_{\text{ref}}, \mathcal{O}_{\text{gen}})}{\max(|\mathcal{N}_{\text{ref}}|, |\mathcal{N}_{\text{gen}}|)}
\label{eq:EDT}
\end{equation}
where $|\mathcal{N}|$ denotes the number of nodes in the tree.

\textbf{Format Compliance Reward} To ensure valid output structure, we define a binary format reward:

\begin{equation}
R_{\text{format}}(\mathcal{O}_{\text{gen}}) = \begin{cases} 
1 & \text{if } \mathcal{O}_{\text{gen}} \text{ follows required schema} \\
0 & \text{otherwise}
\end{cases}
\end{equation}

\textbf{Overall Reward Computation} The final reward combines both components with weighting:
\begin{equation}
\begin{split}
R_{total}(O_{gen}, O_{ref}) &= \lambda R_{struct}(O_{gen}, O_{ref}) \\
&\phantom{=} + (1-\lambda)R_{format}(O_{gen})
\end{split}
\tag{5}
\end{equation}
where $\lambda \in [0,1]$ is a weighting parameter that controls the relative importance of two rewards.

% To curate a high-quality corpus of survey articles, we filtered papers based on title keywords and structural completeness. Additionally, we performed outline deletion and reference completion for each data. Finally, we enriched the references with abstracts retrieved via semantic matching and distilled CoT rationales using the \texttt{DeepSeek-R1} model.

% \subsection{Dataset format}
% Our dataset is a JSONL file organized into four main key fields: "meta", "outline", "cot", and "ref meta". The "meta" field contains paper metadata, including ID, authors, title, and abstract. "outline" provides a list of chapter outlines with their respective IDs, titles, and actually cited references. "ref meta" includes information about all citations, such as keywords, titles, and abstracts. The "cot" field contains the chain-of-thought distilled using \texttt{DeepSeek-R1}.

\begin{table*}[htbp]
\centering
% \footnotesize
\setlength{\tabcolsep}{4pt}
\begin{tabular}{lccccc c c}
\toprule
\multirow{3}{*}{\textbf{Model}} & 
\multicolumn{6}{c}{\textbf{LLM-as-a-Judge $\uparrow$}} & 
\multirow{3}{*}{\textbf{\shortstack{Structural\\Distance}} $\downarrow$} \\
\cmidrule(lr){2-7}
& \textbf{\shortstack{Structure\\Locate}} & 
  \textbf{\shortstack{Structure\\Detail}} & 
  \textbf{\shortstack{Content\\Exclusion}} & 
  \textbf{\shortstack{Content\\Depth}} & 
  \textbf{\shortstack{Pragmatics\\Concise}} & 
  \textbf{\shortstack{Total}} \\
\midrule
Human-written        & 7.80 & 7.21 & 7.93 & 6.00 & 8.29 & 37.23 & 0.00 \\
\midrule
DeepSeek-R1          & 8.09 & 5.31 & 8.15 & 5.75 & 8.85 & 36.15 & 0.48 \\
DeepSeek-V3          & 8.26 & 4.17 & 8.04 & 5.46 & 8.95 & 34.88 & 0.46 \\
GPT-5 Nano           & 6.99 & 5.66 & 6.19 & 5.04 & 7.84 & 31.72 & 0.46 \\
Gemini 2.5 Flash-Lite& 8.01 & 5.68 & 8.18 & 5.84 & 8.72 & 36.43 & 0.52 \\
Qwen3-8B              & 4.97 & 3.71 & 4.47 & 3.65 & 4.62 & 21.42 & 0.59 \\
\rowcolor{LightCyan} Meow-8B-SFT   & 7.79 & 6.75 & 7.11 & 5.34 & 8.11 & 35.10 & 0.43 \\
\rowcolor{LightCyan} Meow-8B-SFT-GRPO  & 8.01 & 6.46 & 7.98 & 5.73 & 8.61 & \textbf{36.79} & \textbf{0.39} \\
\bottomrule
\end{tabular}
\caption{Evaluation results on our self-constructed test set. } % Scores from \textit{LLM-Judge} are better when higher, while \textit{Structural Distance} is better when lower.
\label{tab:ours100}
\end{table*}

\begin{table*}[htbp]
\centering
% \footnotesize
\setlength{\tabcolsep}{4pt}
\begin{tabular}{lccccc c c}
\toprule
\multirow{3}{*}{\textbf{Model}} & 
\multicolumn{6}{c}{\textbf{LLM-as-a-Judge $\uparrow$}} & 
\multirow{3}{*}{\textbf{\shortstack{Structural\\Distance}} $\downarrow$} \\
\cmidrule(lr){2-7}
& \textbf{\shortstack{Structure\\Locate}} & 
  \textbf{\shortstack{Structure\\Detail}} & 
  \textbf{\shortstack{Content\\Exclusion}} & 
  \textbf{\shortstack{Content\\Depth}} & 
  \textbf{\shortstack{Pragmatics\\Concise}} & 
  \textbf{\shortstack{Total}} \\
\midrule
Human-written      & 8.11 & 7.39 & 8.14 & 5.21 & 8.07 & 36.92 &  0.00 \\
\midrule
SurveyX            & 7.37 & 5.30 & 4.60 & 3.90 & 8.43 & 29.60 &  0.44 \\
\rowcolor{LightCyan} Meow-8B-SFT   & 6.56 & 5.83 & 7.33 & 5.28 & 7.83 & 32.83 &  0.53 \\
\rowcolor{LightCyan} Meow-8B-SFT-GRPO  &  7.71  &  6.12  &  7.71  &  5.46  &  8.62  &  \textbf{35.62}   & \textbf{0.42} \\
\bottomrule
\end{tabular}
\caption{Evaluation results on the SurveyX test set. } % Scores from \textit{LLM-Judge} are better when higher, while \textit{Structural Distance} is better when lower.
\label{tab:surveyx}
\end{table*}

\section{Experiments}

\subsection{Datasets}

We divided the raw dataset into an SFT training set, an RL training set, and a test set. 
% Based on our outline quality criteria, we selected data with total scores above 39 for training. To prevent overlong inputs and improve training efficiency, we limited the input length to 32k tokens and the ground truth to 10k tokens for the SFT set, and to 20k and 10k tokens, respectively, for the RL set. Ultimately, we constructed an SFT training set comprising 870 articles and an RL training set containing 180 articles. 
The evaluation of survey outlines is inherently challenging due to the lack of a standardized benchmark. 
The primary test set is constructed by selecting 100 survey articles with \texttt{update\_date} after 2025. 
This design minimizes the possibility that these papers were seen during model pretraining, thereby ensuring evaluation fairness. 
Broad domain coverage is further guaranteed by including articles from all major academic disciplines. 
In addition, the 13 papers used in the SurveyX examples\footnote{https://github.com/IAAR-Shanghai/SurveyX/tree/main/examples} are incorporated as an auxiliary test set. 
To avoid data leakage, all articles in the evaluation set are removed from the training corpus. All training and test sets are released in huggingface datasets\footnote{https://huggingface.co/datasets/haajimi/Meow}.

\subsection{Evaluation Metrics}

\textbf{LLM-as-a-Judge} LLMs are employed as judges to evaluate whether an outline effectively supports two reader needs: 
(1) quick access to specific knowledge details and related theories in the field, and 
(2) deeper understanding of domain insights, capturing trends and directions to guide follow-up research.  
Accordingly, the evaluation is organized along three dimensions: \textit{structure}, \textit{content}, and \textit{pragmatics}, each further operationalized into five concrete criteria:

\textbf{(1) Structure Locate}: Adherence to conventional organizational frameworks (e.g., IMRaD) for efficient information retrieval.
\textbf{(2) Structure Detail}: Appropriate space allocation based on topic importance and complexity, emphasizing core content over secondary details.
\textbf{(3) Content Exclusion}: Clear boundaries between same-level sections to avoid redundant overlap.
\textbf{(4) Content Depth}: Integration of diverse reasoning structures (e.g., causal links, theory-to-application) and progressive logical chains across sections.
\textbf{(5) Pragmatics Concise}: Concise yet descriptive section titles without overly broad expressions.

Scores from LLM-Judge are reported on a 0–10 scale, with higher scores indicating better performance. We provide evaluation scripts on GitHub\footnote{https://github.com/cedricshan/Survey-Outline-Evaluation-Benckmark}.

\textbf{Structural Distance} We use the tree edit distance as structural distance to measure the difference between a generated outline and its human-written counterpart. 
A lower distance indicates a structure closer to the human reference.

\subsection{Baselines}

We select several representative LLMs as baselines, including \texttt{DeepSeek-R1}, \texttt{DeepSeek-V3},  \texttt{GPT-5 Nano}, and \texttt{Gemini 2.5 Flash-Lite}. 
All models are accessed via their official APIs with default temperature and max token settings. 
In addition, we include \texttt{SurveyX}~\cite{liang2025surveyxacademicsurveyautomation} as a representative work from the academic community for comparison, together with human-written as a ground truth.

\subsection{Main Results}

We train our model based on \texttt{Qwen3-8B}\footnote{https://huggingface.co/Qwen/Qwen3-8B}. Our models consistently achieve strong performance across test sets. On our self-constructed test set (Table \ref{tab:ours100}), Meow-8B-SFT reaches an LLM-Judge score of 35.10, while Meow-8B-SFT-GRPO improves to 36.79 and attains the lowest Structural Distance of 0.39, indicating that reinforcement learning enhances structural fidelity. On the SurveyX test set (Table \ref{tab:surveyx}), Meow-8B-SFT achieves 32.83 compared to the baseline 29.60, and Meow-8B-SFT-GRPO further improves to 35.62, demonstrating consistent advantages in structural coherence and bibliographic accuracy. Representative LLMs remain unable to match the writing quality of high-quality human-crafted survey outlines, underscoring that end-to-end survey outline generation continues to pose significant challenges. Our models have achieved performance that most closely approximates the standard of expert-written outlines.

Notably, although the GRPO process aligns the model with human behavior only at the outline-structure level, it yields complementary gains across other metrics. This indicates that emulating the structured reasoning paradigms of expert researchers not only enhances structural fidelity but also implicitly improves overall writing quality, making the generated survey outlines more consistent with human research logic and academic conventions.

\section{Conclusion}
This paper presents \textbf{Meow}, a metadata-driven framework combining CoT cold-start with reinforcement learning for survey outline generation. Experiments demonstrate superior structural fidelity and stylistic coherence, providing a scalable pathway toward automated survey generation.

\clearpage
\bibliographystyle{IEEEbib}
\bibliography{strings,refs}

\end{document}